\documentclass[10pt,twocolumn,letterpaper]{article}

\usepackage{cvpr}              

\usepackage{microtype}

\renewcommand{\paragraph}[1]{\vspace{.5em}\noindent\textbf{#1.}}

\definecolor{cvprblue}{rgb}{0.21,0.49,0.74}
\usepackage[pagebackref,breaklinks,colorlinks,allcolors=cvprblue]{hyperref}
\usepackage{multirow}
\usepackage{booktabs}
\usepackage{graphicx}
\usepackage{subcaption}
\usepackage{float}
\usepackage{caption}
\usepackage{booktabs}
\usepackage{adjustbox}
\usepackage{makecell,siunitx}
\usepackage{comment}

\def\paperID{11} 
\def\confName{CVPR}
\def\confYear{2026}

\title{Rethinking Satellite Image Restoration for Onboard AI: A Lightweight Learning-Based Approach}

\author{
Adrien Dorise$^{1,2}$, Marjorie Bellizzi$^{1}$, Omar Hlimi$^{1}$\\[0.5em]
$^{1}$Institut de Recherche Technologique Saint Exupéry, Toulouse, France\\
$^{2}$Centre national d'études spatiales, Toulouse, France\\[0.5em]
{\tt\small adrien.dorise@cnes.fr,}\\
{\tt\small \{marjorie.bellizzi, omar.hlimi\}@irt-saintexupery.com}
}

\begin{document}
\maketitle

\newcommand{\FigDiorExample}{
\begin{figure}[t]
\centering
\label{fig:dior_restored}

\begin{adjustbox}{max width=\linewidth}
\begin{tabular}{c@{\hspace{4pt}}c@{\hspace{4pt}}c}
\toprule
\textbf{Deg-High} & \textbf{Reference} & \textbf{ConvBEERS} \\
\midrule

\includegraphics[width=0.32\linewidth]{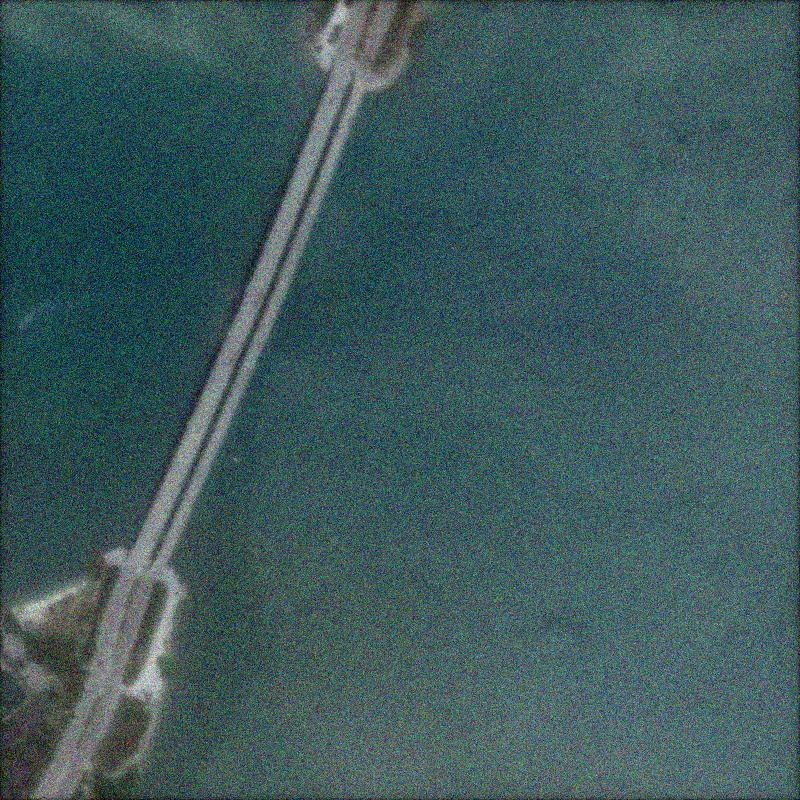}
& \includegraphics[width=0.32\linewidth]{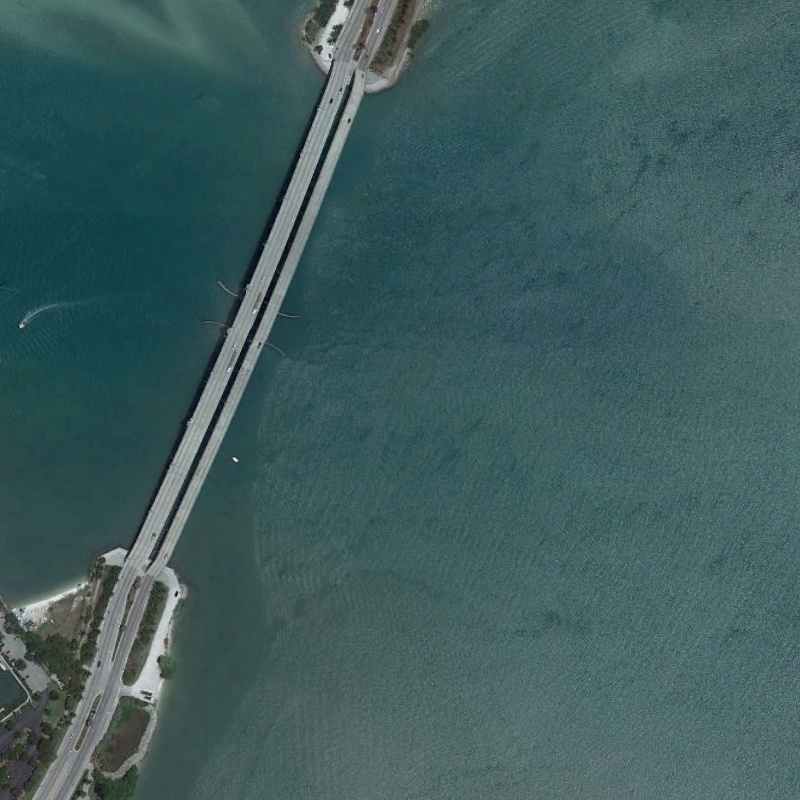}
& \includegraphics[width=0.32\linewidth]{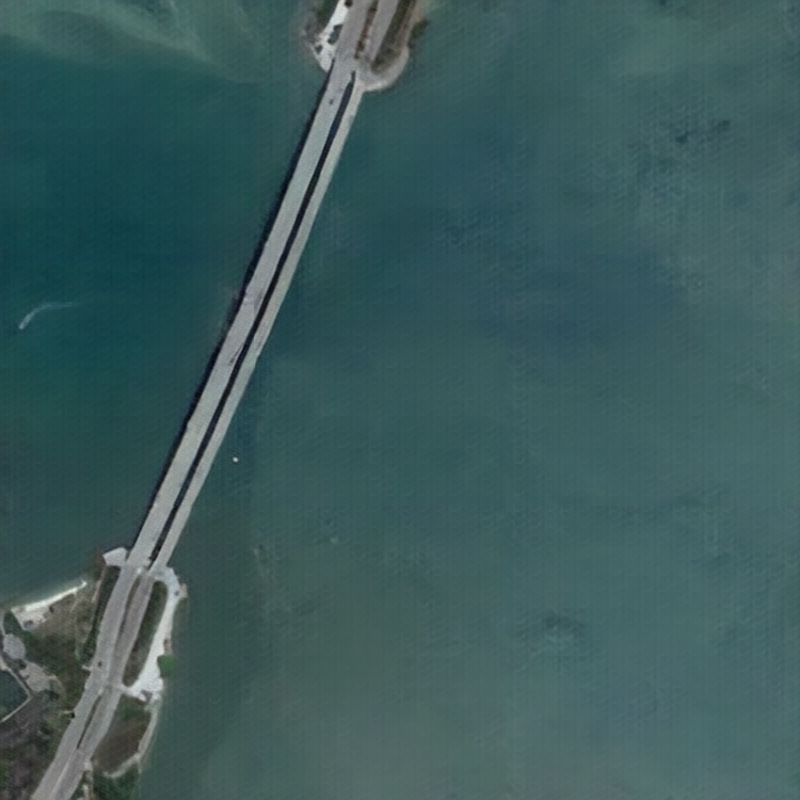} \\

\includegraphics[width=0.32\linewidth]{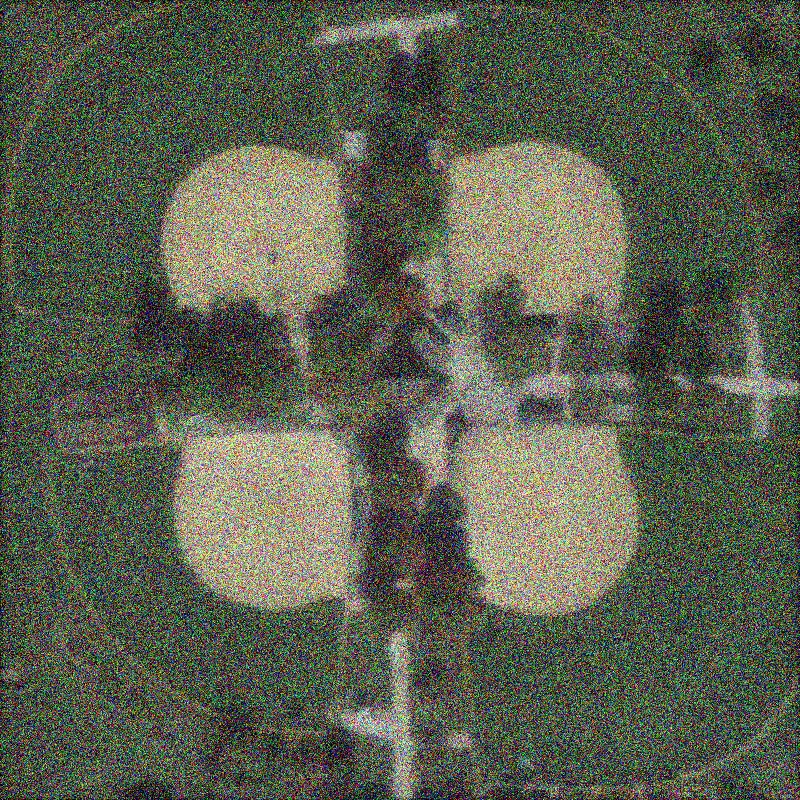}
& \includegraphics[width=0.32\linewidth]{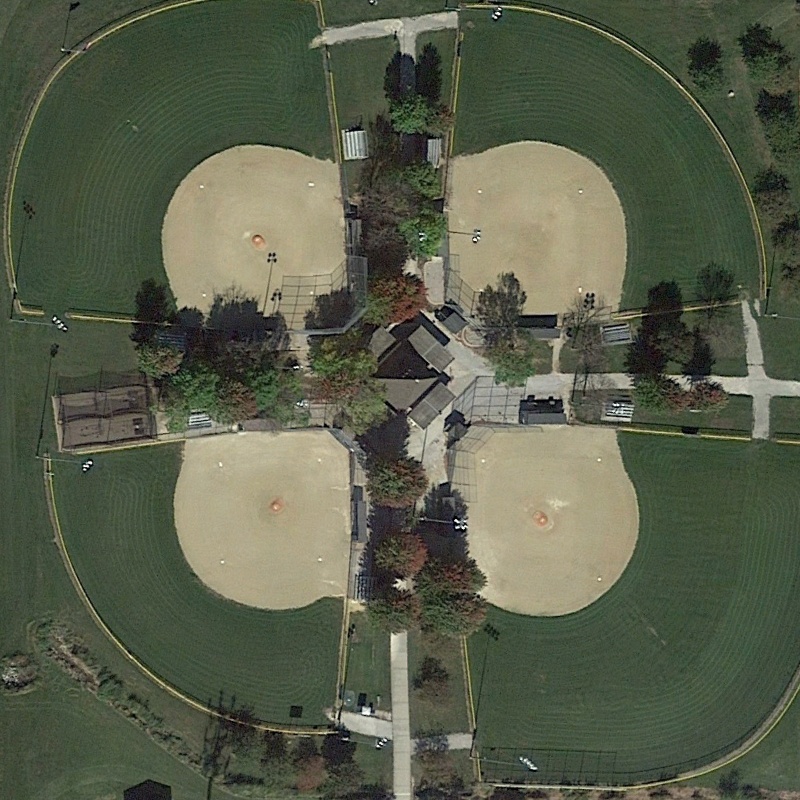}
& \includegraphics[width=0.32\linewidth]{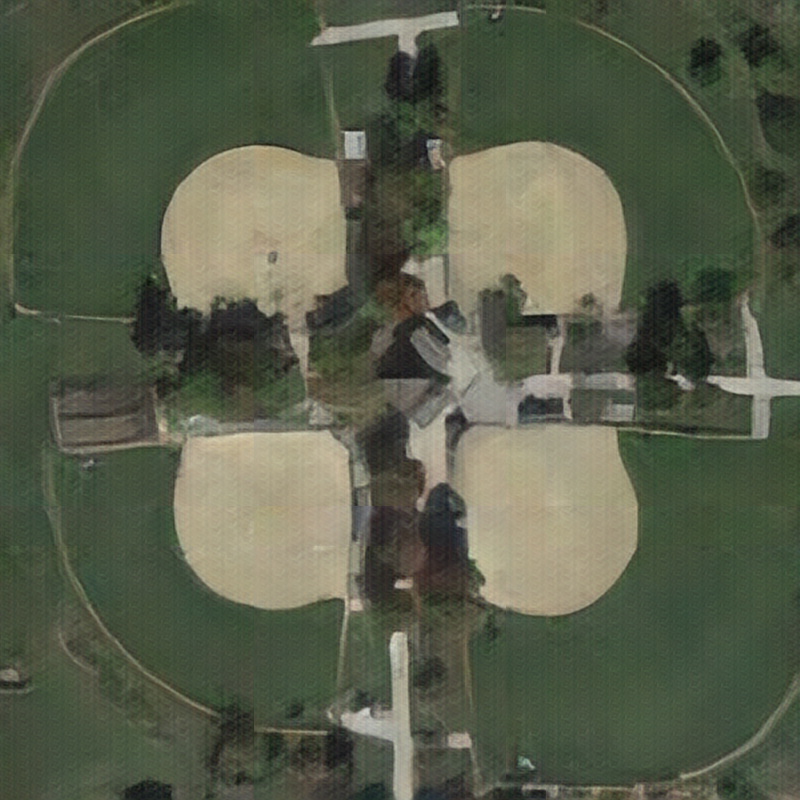} \\

\includegraphics[width=0.32\linewidth]{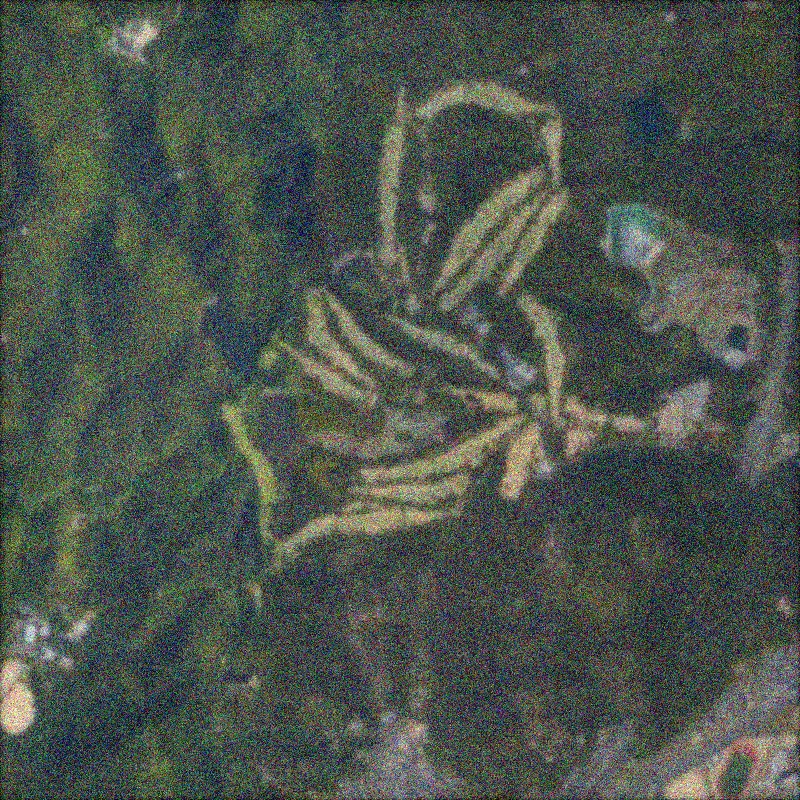}
& \includegraphics[width=0.32\linewidth]{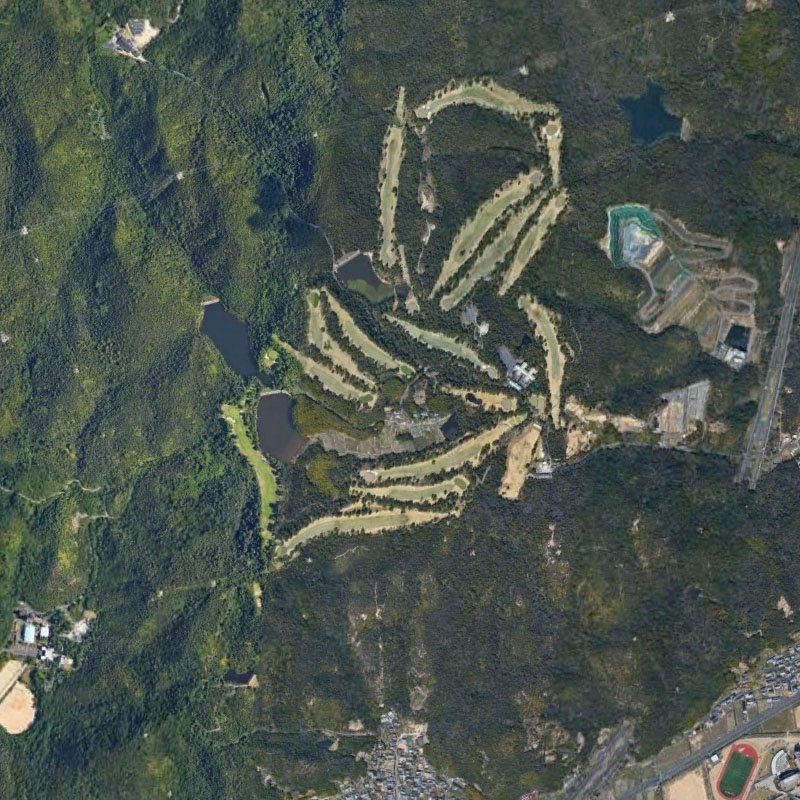}
& \includegraphics[width=0.32\linewidth]{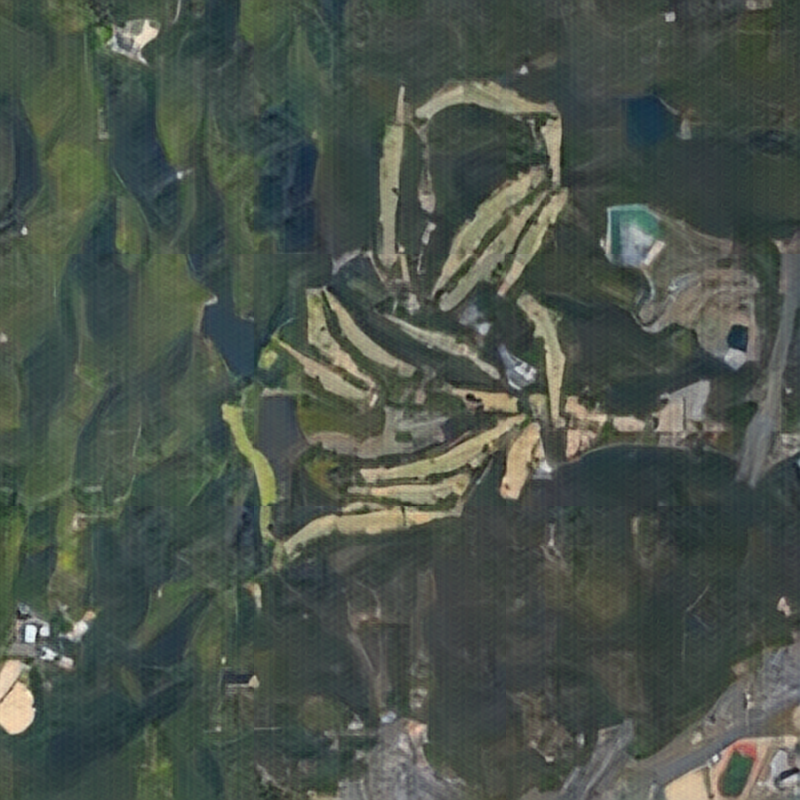} \\

\bottomrule
\end{tabular}
\end{adjustbox}

\caption{Qualitative comparison of restored RGB images with ConvBEERS}
\end{figure}
}

\newcommand{\FigModelTraining}{
\begin{figure}[htbp]
    \centering
    \begin{subfigure}{0.45\textwidth}
        \centering
        \includegraphics[width=\linewidth]{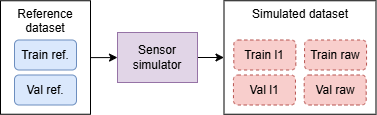}
        \caption{Dataset simulation pipeline}
        \label{fig:dataset_pipeline}
    \end{subfigure}
    \hfill
    \begin{subfigure}{0.45\textwidth}
        \centering
        \includegraphics[width=\linewidth]{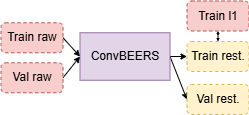}
        \caption{ConvBEERS training}
        \label{fig:Convbeers_training}
    \end{subfigure}
    \caption{ConvBEERS training pipeline. Images are firstly sent to our sensor simulator before being used to train the model}
    \label{fig:main}
\end{figure}
}

\newcommand{\FigDiorSetup}{
\begin{figure}[htbp]
    \centering
    \begin{subfigure}{0.45\textwidth}
        \centering
        \includegraphics[width=\linewidth]{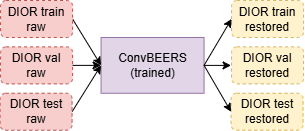}
        \caption{Dataset simulation pipeline}
        \label{fig:dataset_pipeline}
    \end{subfigure}
    \hfill
    \begin{subfigure}{0.45\textwidth}
        \centering
        \includegraphics[width=\linewidth]{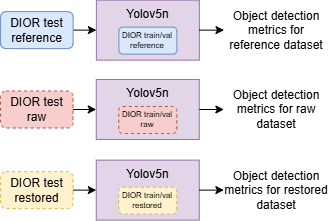}
        \caption{Object detection pipeline}
        \label{fig:Convbeers_training}
    \end{subfigure}
    \caption{Object detection pipeline. }
    \label{fig:main}
\end{figure}
}

\newcommand{\FigDiorPipeline}{
\begin{figure*}[t]
    \centering
    \includegraphics[width=0.98\textwidth]{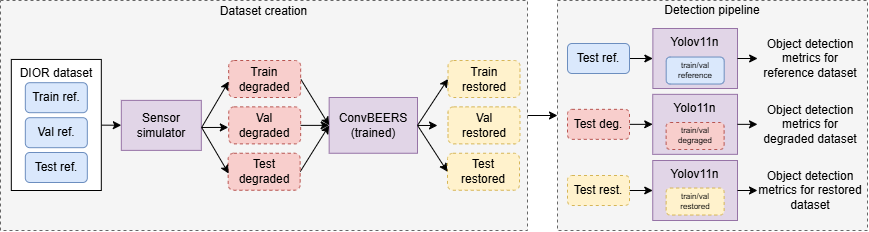}
    \caption{\textbf{Object detection evaluation pipeline}. The simulated degraded and restored datasets are created from the DIOR dataset using both our sensor simulation framework and ConvBEERS. Then, three YOLOv11n models are trained independently on (1) reference images, (2) degraded images, and (3) restored images using the training and validation sets. Their performance is compared on the test sets to evaluate the impact of restoration on a detection task.}
    \label{fig:object_detection_pipeline}
\end{figure*}
}

\begin{abstract}
Satellite image restoration aims to improve image quality by compensating for degradations (e.g., noise and blur) introduced by the imaging system and acquisition conditions. As a fundamental preprocessing step, restoration directly impacts both ground-based product generation and emerging onboard AI applications. Traditional restoration pipelines based on sequential physical models are computationally intensive and slow, making them unsuitable for onboard environments. In this paper, we introduce \mbox{ConvBEERS}: a Convolutional Board-ready Embedded and Efficient Restoration model for Space to investigate whether a light and non-generative residual convolutional network, trained on simulated satellite data, can match or surpass a traditional ground-processing restoration pipeline across multiple operating conditions. 

Experiments conducted on simulated datasets and real Pleiades-HR imagery demonstrate that the proposed approach achieves competitive image quality, with a +6.9dB PSNR improvement. Evaluation on a downstream object detection task demonstrates that restoration significantly improves performance, with up to +5.1\% mAP@50. In addition, successful deployment on a Xilinx Versal VCK190 FPGA validates its practical feasibility for satellite onboard processing, with a $\sim$41$\times$ reduction in latency compared to the traditional pipeline. These results demonstrate the relevance of using lightweight CNNs to achieve competitive restoration quality while addressing real-world constraints in spaceborne systems.
\end{abstract}

\section{Introduction}
\label{sec:intro}

High-resolution optical satellite imagery is acquired at the raw level onboard, where sensor optical limitations and acquisition conditions, including Modulation Transfer Function (MTF) effects and Signal-to-Noise Ratio (SNR), inherently impact image quality. Image restoration aims to improve image quality by jointly addressing optical blur, sensor noise, and acquisition-related effects.
It therefore constitutes a prerequisite preprocessing step for any higher-level exploitation of satellite imagery.
This is true not only for traditional ground-based workflows but also for emerging onboard processing pipelines, where early image quality enhancement can directly benefit downstream analytical tasks.

Traditional restoration approaches are typically based on sequential physical models that combine deconvolution and denoising stages \cite{pleiades_restoration, carlavan_restoration, NL_bayes_for_restoration}. While robust and well-established, these ground-based pipelines are computationally demanding, with iterative optimisation and multiple buffers. This results in significant memory usage and processing latency, which severely limit their suitability for onboard implementation. 

The increasing demand for low-latency applications, including disaster response \cite{ onboard_oil_spill_detection, irma_imagini}, maritime surveillance \cite{maritime_surveillance_cyprus, vessel_detection_yolo11_sentinel2} and autonomous onboard decision-making \cite{pose_estimation_IRT}, motivates a shift toward more efficient processing strategies. In this context, recent studies evaluate the feasibility of learning-based approaches for remote sensing applications \cite{deep_priors_for_restoration,thesis_restoration_AI,pseudo_generative_for_restoration, AI_for_pleiades}, offering a promising alternative for ground and board processing. However, limited access to raw imagery often limits their use for onboard applications \cite{pyraws_maritime1, pyraws_maritime2, pyraws_thraws}.

In this work, we investigate whether a light, non-generative residual convolutional neural network trained exclusively on physically realistic simulated satellite data can achieve restoration performance comparable to that of a traditional ground-processing pipeline when applied to both simulated and real imagery. To do so, we introduce \mbox{ConvBEERS}, a Convolutional Bord-ready Embedded and Efficient Restoration model for Space. Our approach proposes a novel training strategy on physics-driven simulated images. Beyond image quality assessment, we also evaluate the impact of such restoration on AI-based remote sensing applications, specifically object detection. Finally, we analyse the onboard performance of our architecture by deploying it on a Xilinx Versal VCK190 FPGA, demonstrating its suitability under the memory, latency, and power constraints typical of spaceborne systems.

In this work, we investigate whether a light, non-generative residual convolutional neural network, EDSR \cite{EDSR}, trained exclusively on physically realistic simulated satellite data, can achieve restoration performance comparable to that of a traditional ground-processing pipeline when applied to both simulated and real Pleiades imagery.



\section{Background and related work}
\label{sec:RelatedWork}

Remote sensing products are distributed according to \textit{processing levels}, which reflect the degree of radiometric and geometric correction applied to the raw sensor measurements. The first available product is generally a \mbox{Level-1} image, obtained after radiometric calibration, geometric correction, co-registration, and orthorectification of the raw acquisition \cite{sentinel2_products}. Here, we concentrate on radiometric enhancement.

\paragraph{Traditional restoration}
Classical radiometric restoration pipelines typically combine deconvolution and denoising stages. In very-high-resolution optical systems such as Pléiades, restoration relies on an instrument-driven deconvolution step followed by advanced denoising methods such as NL-Bayes \cite{NL_bayes_for_restoration, pleiades_restoration, carlavan_restoration}. These methods rely on explicit physical modelling of the imaging system (system point spread function (PSF) and noise statistics), enabling improved sharpness while controlling noise amplification. Although operationally validated \cite{pleiades_restoration,fusion_pleiades}, such pipelines involve sequential processing steps and iterative operations. Consequently, they are primarily suited for ground-based processing and remain difficult to adapt to onboard or real-time deployment. Moreover, because they are tailored to specific optical system characteristics, generalisation to other systems is challenging.



\paragraph{AI-based restoration}
In parallel, learning-based approaches have emerged for raw image restoration \cite{deep_priors_for_restoration,VAE_for_restauration, AI_for_pleiades}. CNN-based architectures are commonly employed due to their efficient convolutional operations and ability to model local spatial structures \cite{wavelet_for_noise_removal, deep_connected_cnn_for_restoration}. Their residual layers facilitate stable optimisation while maintaining moderate computational complexity. Generative strategies have also been explored. GAN-based methods optimise perceptual and adversarial objectives to enhance visual sharpness \cite{SRGAN}, but may synthesise high-frequency components and are prone to creating artefacts \cite{perceptual_fidelite}. In remote sensing applications, where geometric accuracy and radiometric consistency are critical, such hallucinated structures can affect quantitative reliability. Transformer-based models and diffusion models have demonstrated competitive performance in image restoration by capturing long-range dependencies through self-attention mechanisms \cite{transformer_restoration, hybrid_transfomer_for_image_restoration, diffusion23}, but rely on large backbone architectures and computationally intensive inference procedures, leading to significant computational and memory demands that are incompatible with resource-constrained onboard deployment.



\paragraph{Onboard restoration in remote sensing}
Ongoing work studies the feasibility of light-based restoration models \cite{lightweight_cnn_restoration_for_aerial_low_visibility}, but despite the rapid progress of AI-based restoration methods, onboard image processing pipelines in remote sensing still largely replicate traditional ground-based approaches. A few experimental demonstrations have explored learning-based enhancement in orbit. Notably, the OPS-SAT mission reported a generative AI-based image denoising experiment deployed on board, demonstrating the feasibility of neural restoration under real-space constraints \cite{gan_restoration_onboard_opsat}. More recently, scalable neural pushbroom architectures for real-time denoising have been implemented on NVIDIA Jetson-class embedded hardware, highlighting the potential of lightweight CNN designs for edge inference in space environments \cite{jetson_restoration_AI}. Other recent approaches, such as PyRaws \cite{pyraws_thraws}, primarily focus on restoring the geometric aspects of raw satellite data through coarse-coregistration.

\paragraph{Positioning of this work}
Previous work have demonstrated the feasibility of deploying learning-based image enhancement onboard satellites. However, these works typically focus either on specific enhancement tasks or on proof-of-concept deployment, without systematic comparison to operational restoration pipelines or evaluation within a complete application chain. In contrast, this work investigates whether a lightweight, non-generative CNN trained exclusively on physics-based simulated data can match the performance of traditional model-based restoration, while satisfying FPGA-onboard constraints. Beyond image quality assessment, we evaluate its impact on an end-to-end use case, thus positioning restoration as a task-oriented preprocessing module for operational spaceborne AI systems.

\section{Proposed method}
\label{sec:proposed_methods}


\begin{figure}[htbp]
    \centering
    \includegraphics[width=\linewidth]{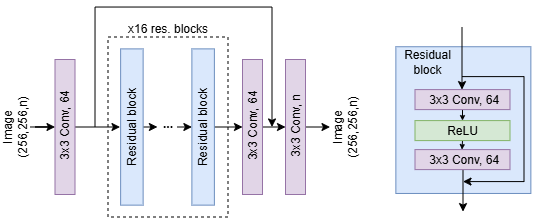}
    \caption{ConvBEERS architecture. It is composed of 3x3 size-preserving convolution layers (stride=1, padding=1) and 16 residual blocks without batch normalisation for a total of 1.2M trainable parameters.}
    \label{fig:ConvBEERS}
\end{figure}

\subsection{A light residual network for restoration}
\label{sec:ConvBEERS}
Residual connections in convolutional neural networks greatly improve the flow of gradient, and allow for deeper architectures \cite{resnet}. This was later successfully adapted into super-resolution models \cite{SRGAN, EDSR}. Image super-resolution can be formulated as an inverse image restoration problem where the degradation operator consists of blurring and downsampling \cite{deep_cnn_denoiser_for_image_restoration, deep_cnn_for_image_denoising}. 

With that in mind, we built upon the residual layers without batch normalisation design of Enhanced Deep Super-Resolution (EDSR) \cite{EDSR} while adapting it for the restoration by removing the upsample layers. 

We propose \mbox{ConvBEERS}, a Convolutional Board-ready Embedded and Efficient Restoration model for Space, as a light, non-generative learning-based restoration technique. \mbox{ConvBEERS} is a fully convolutional architecture composed of a sequence of 16 residual blocks without batch normalisation and followed by a reconstruction layer, as shown in Fig.\ref{fig:ConvBEERS}.
The absence of adversarial training and generative components ensures stable optimisation and avoids hallucination of artificial structures, which is critical for operational satellite imagery. The residual design encourages the model to learn how to correct the degradations with respect to the input image, and the absence of batch normalisation helps preserve feature magnitude and overall image consistency. Training is performed in a fully supervised manner using simulated image pairs (\textit{Sim-Degraded-Variable / Sim-Reference-Fixed}) only, to assess the intrinsic generalisation capability of the model.

In addition, \mbox{ConvBEERS} exhibits strong portability properties for onboard deployment. As a fully convolutional network, it is resolution-agnostic and can process large images through tiling without architectural changes. The absence of batch normalisation simplifies inference and facilitates quantised or fixed-point implementations on embedded hardware. Moreover, the model does not require scene-dependent parameter tuning at inference time, enabling straightforward integration across sensors and evolving operating conditions.

\subsection{Training objective}
To balance spatial fidelity, perceptual realism, and spectral consistency, we employ a composite loss function combining pixel-domain, perceptual, and frequency-domain supervision:
\begin{equation}
\mathcal{L}_{total} =
\lambda_{1} \mathcal{L}_{L1} +
\lambda_{p} \mathcal{L}_{LPIPS} +
\lambda_{f} \mathcal{L}_{FFT},
\end{equation}
where $\lambda_{1}$, $\lambda_{p}$, and $\lambda_{f}$ denote weighting coefficients. In this work, experiments were performed with $\lambda_{1}=1.0$, $\lambda_{p}=0.5$ and $\lambda_{f}=0.1$

\paragraph{Pixel reconstruction loss}
We adopt the $\ell_1$ loss to enforce pixel-wise consistency:
\begin{equation}
\mathcal{L}_{L1} = \| x - \hat{x} \|_1,
\end{equation}
which has been shown to better preserve sharp structures than $\ell_2$ in super-resolution and restoration tasks \cite{EDSR}.

\paragraph{Perceptual loss}
To encourage visually meaningful reconstructions, we incorporate LPIPS \cite{lpips,loss_function_for_SR}, which measures similarity in deep feature space:
\begin{equation}
\mathcal{L}_{LPIPS} =
\sum_{l} w_l \| \phi_l(x) - \phi_l(\hat{x}) \|_2^2,
\end{equation}
where $\phi_l(\cdot)$ denotes feature activations extracted from a pretrained network. Perceptual supervision has proven effective for restoration and super-resolution, preserving semantic structure and fine details.

\paragraph{Frequency-domain loss}
Since degradation processes primarily attenuate high-frequency components, we additionally enforce spectral consistency by minimising the $\ell_1$ distance between Fourier representations:
\begin{equation}
\mathcal{L}_{FFT} =
\left\| \mathcal{F}(x) - \mathcal{F}(\hat{x}) \right\|_1,
\end{equation}
where $\mathcal{F}$ denotes the 2D Fourier transform. Frequency-domain supervision has been shown to improve reconstruction of fine textures and structural details in image restoration \cite{focal_frequency_loss_for_image_reconstruction,frequency_for_sr}.

\subsection{Sensor simulation}
\label{sec:sensor_simulation}
Learning-based image restoration requires paired data, where a simulated degraded image is paired with a simulated reference image used for training. In operational satellite imaging, only raw-level images are accessible, while ground-truth level-1 (L1) images are inherently unavailable. Consequently, supervised training cannot rely on real raw/L1 pairs without biasing the model toward reproducing a specific processing chain.
To overcome this limitation, we construct a physics-based sensor simulator that approximates realistic satellite imaging conditions while maintaining full control over the simulation parameters. From there, we use the sensor simulator to degrade Level-1 products, creating simulated Level-1 and simulated raw imagery while limiting the impact of the original Level-1 restoration on the image. By doing so, we ensure better control over our experiment parameters.

\paragraph{Spatial resolution}
Spatial blur is modelled through a parametric MTF combining optical and sensor effects:
\begin{equation}
\mathrm{MTF}(f_x,f_y) 
= \exp(-\gamma f_r) 
\cdot \mathrm{sinc}(f_x) 
\cdot \mathrm{sinc}(f_y),
\end{equation}
where
\[
f_r = \sqrt{f_x^2 + f_y^2}
\]
and $\gamma$ is calibrated from a Nyquist value $MTF_{Nyq}$.  
The corresponding Point Spread Function (PSF) is obtained as
\begin{equation}
\mathrm{PSF} = \mathcal{F}^{-1}\{\mathrm{MTF}\},
\end{equation}
and each spectral band is convolved with the PSF.

\paragraph{Ground Sampling Distance GSD}
To simulate a coarser GSD, the blurred image is downsampled by a factor $r$ (the oversampling ratio). Prior to subsampling, an anti-aliasing Gaussian filter is applied. The image is then sampled every $r$ pixels:
\begin{equation}
I_{\text{GSD}}(x,y) = I_{\text{blur}}(x_0 + r x,\; y_0 + r y),
\end{equation}
where $(x_0,y_0)$ denotes a centred offset ensuring no spatial shift. This procedure simulates both resolution loss and enlargement of the pixel footprint.

\paragraph{Signal-dependent noise model}
\label{sec:signal_dependent_noise}
Radiometric degradation is modelled using signal-dependent Gaussian noise centered on 0 and with standard deviation:
\begin{equation}
\sigma^2(L) = \alpha L + \beta,
\end{equation}
where $L$ is the luminance. The parameters $\alpha$ and $\beta$ are estimated from two reference luminance--SNR pairs $(L_0,\mathrm{SNR}_0)$ and $(L_1,\mathrm{SNR}_1)$:
\begin{equation}
\sigma_i^2 = \left(\frac{L_i}{\mathrm{SNR}_i}\right)^2,
\qquad
\sigma_i^2 = \alpha L_i + \beta.
\end{equation}
Noise is sampled as
\begin{equation}
\sigma(L) = \sqrt{\alpha L + \beta},
\end{equation}
and added independently to each pixel.

By varying $MTF_{Nyq}$, the downsampling factor $r$, and the reference SNR values, multiple degradation levels are generated, enabling systematic evaluation under controlled and realistic imaging conditions. In our experiments, \mbox{ConvBEERS} is trained solely on simulated images.

\section{Experimental Setup}
\label{sec:materials}


\subsection{OpenAerialMap dataset}
\label{sec:datasets}
Our dataset is constructed from high-resolution aerial RGB imagery obtained from OpenAerialMap \cite{openaerialmap}, with an initial GSD of 10\,cm. This resolution provides sufficient spatial margin to realistically simulate satellite acquisition conditions by degrading the data toward the targeted GSD, MTF, and SNR operating ranges. The dataset comprises a wide variety of landscape types (urban, suburban, rural, agricultural, forested, coastal, and mountainous), with an emphasis on urban environments characterised by high spatial frequencies, sharp edges, and dense structural details. Such scenes are particularly challenging for image restoration and constitute a robust evaluation of the model’s ability to recover fine details without introducing artefacts such as ringing or oversmoothing.

From there, these aerial RGB images are converted to panchromatic imagery and degraded using our physics-based sensor simulator pipeline (see Sec.\ref{sec:sensor_simulation}), which models the key characteristics of the target system (GSD, MTF, and SNR). This process generates paired simulated raw and reference images under controlled, reproducible degradation settings.
Several configurations are produced at a target GSD of 50\,cm and stored as 12-bit panchromatic TIFF patches. Their specificities are disclosed in Tab.~\ref{tab:datasets}. 

Two types of datasets are constructed. The first type, referred to as \textit{Sim-Degraded-Fixed / Sim-Reference-Fixed}, corresponds to a fixed-degradation setting defined at a nominal operating point representative of typical acquisition conditions. The second type consists of variable-degradation datasets, obtained by sampling MTF and SNR values within realistic ranges around this nominal configuration. Introducing such variability in the degradation process encourages the model to learn to handle diverse imaging conditions. It improves robustness while reducing the risk of overfitting to a specific system configuration.


\begin{table}[t]
\centering
\caption{Summary of simulated and real datasets used in this work.}
\label{tab:datasets}
\footnotesize
\begin{tabular}{lccc}
\hline
\textbf{Dataset} & \textbf{MTF} & \textbf{SNR @ $L_0$ / $L_1$} & \textbf{\# Patches} \\
\hline
Sim-Degraded-Variable   & 3--7\%     & 50 $\pm$ 40 / 110 $\pm$ 40 & 196\,128 / 36 \\
Sim-Degraded-Fixed      & 7\%        & 50 / 110                        & 196\,128 / 36 \\
Sim-Reference-Fixed      & 25\%       & 80 / 170                        & 196\,128 / 36 \\
Real-Raw-Pleiades  & $\sim$7\%  & $\sim$50 / $\sim$110            & -- / 19 \\
\hline
\end{tabular}
\vspace{1mm}

\footnotesize
SNR values (in dB) are defined at reference luminance levels $L_0 = 25$~W/m$^2$/sr/$\mu$m and $L_1 = 100$~W/m$^2$/sr/$\mu$m. Number of patches is reported as \textit{Train / Test}. Training patches are
$128 \times 128$ pixels, while test patches are $1500 \times 1500$ pixels.
\end{table}



\subsection{Metrics}
\label{sec:Metrics}

Restoration performance is evaluated through two complementary perspectives: reference-based image-similarity metrics and physically grounded sensor-level performance indicators, along with qualitative visual inspection.

\paragraph{Full-reference image quality metrics}

When reference images are available, restoration performance is evaluated using classical full-reference image quality metrics, including Peak Signal-to-Noise Ratio (PSNR) and Structural Similarity Index (SSIM). These metrics quantify pixel-level fidelity and structural consistency between restored images and their reference counterparts. 
Reported values are computed over the entire simulated test set, composed of independent images taken from different localisations, and averaged across all test patches to ensure statistically representative evaluation.
Although widely adopted, these metrics may exhibit limited correlation with perceived visual quality, particularly in the presence of blur–noise trade-offs. To better capture perceptual and structural aspects of restoration, we additionally report learned perceptual metrics: LPIPS (Learned Perceptual Image Patch Similarity) and DISTS (Deep Image Structure and Texture Similarity). These feature-based metrics rely on deep neural representations and have demonstrated improved sensitivity to texture reconstruction, edge preservation, and structural distortions.


\paragraph{Physical image quality metrics}

Beyond similarity-based measures, we estimate physical image quality metrics directly related to imaging system performance. In particular, MTF and the SNR are measured on restored images and compared with their input counterparts.

The MTF is estimated using a slanted-edge method implemented in the \textit{MTF Estimator} plugin for QGIS \cite{Gil_MTFEstimator_QGIS_2025}. For each image, measurements are performed on three manually selected edge regions, and the reported value is the average of these regions, ensuring a robust and representative estimate of spatial resolution. The SNR is evaluated over 13 homogeneous regions using the variance-based radiometric approach described by Liu et al. \cite{Liu1999SNR}, following the signal-dependent noise model introduced in Sec.~\ref{sec:signal_dependent_noise}.


These indicators provide a physically interpretable assessment of blur compensation and noise amplification, ensuring consistency with operational imaging specifications. 

\section{Experimental results and embedded system performance analysis}
\label{sec:results}


\subsection{Benchmark on OpenAerialMap}
We evaluate \mbox{ConvBEERS} on the OpenAerialMap dataset and compare its performance to the Pléiades-HR restoration, which serves as a baseline \cite{pleiades_restoration,fusion_pleiades}. The model is trained for 30 epochs on the training subset of \textit{Sim-Degraded-Variable / Sim-Reference-Fixed}, with the degraded set serving as input, and the reference set serving as target (see Sec.~\ref{sec:sensor_simulation}). Both approaches are evaluated on the fixed-degradation simulated test set (\textit{Sim-Degraded-Fixed / Sim-Reference-Fixed}) using the metrics described in Sec.~\ref{sec:Metrics}. 

Results are reported in Tab.~\ref{tab:quality_simulated} and Tab.~\ref{tab:runtime_simulated}. We see that \mbox{ConvBEERS} outperforms the traditional pipeline, with a +6.9 dB PSNR gain, higher SSIM, and a markedly lower LPIPS score. The restored MTF reaches 0.20 while preserving comparable SNR, confirming effective sharpness-blur compensation without significant noise amplification. MTF estimation is not reported for the traditional pipeline since the slanted-edge method did not converge reliably due to nonlinear ringing effects introduced by the deconvolution and denoising stages.

\begin{figure}[htbp]
    \centering
    
    \begin{subfigure}[b]{0.48\linewidth}
        \centering
        \includegraphics[width=\linewidth]{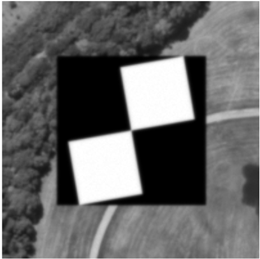}
        \caption{Simulated degraded}
        \label{fig:sub1}
    \end{subfigure}
    \hfill
    \begin{subfigure}[b]{0.48\linewidth}
        \centering
        \includegraphics[width=\linewidth]{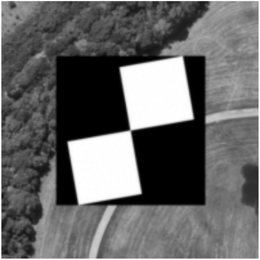}
        \caption{Simulated reference}
        \label{fig:sub2}
    \end{subfigure}

    \vspace{2mm}

    \begin{subfigure}[b]{0.48\linewidth}
        \centering
        \includegraphics[width=\linewidth]{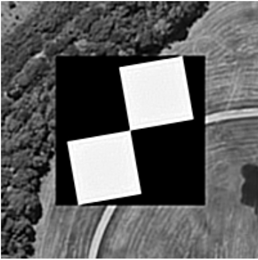}
        \caption{Traditional}
        \label{fig:sub3}
    \end{subfigure}
    \hfill
    \begin{subfigure}[b]{0.48\linewidth}
        \centering
        \includegraphics[width=\linewidth]{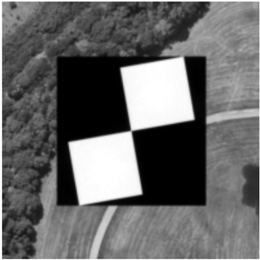}
        \caption{ConvBEERS}
        \label{fig:sub4}
    \end{subfigure}

    \caption{Image restoration on a simulated degraded image with embedded mire.}
    \label{fig:mire}
\end{figure}

In addition to quantitative metrics, a qualitative visual analysis is performed, focusing on edge sharpness, texture preservation, residual noise, and the absence of non-physical artefacts. A synthetic resolution pattern embedded in simulated scenes (Fig.~\ref{fig:mire}) provides a controlled and repeatable assessment of high-frequency reconstruction. Visual observations align with the quantitative results, confirming an effective balance between smoothing and sharpness restoration, with good robustness to noise.

\begin{table}[t]
\centering
\small
\caption{Full-reference image quality metrics on OpenAerialDataset.}
\label{tab:quality_simulated}
\begin{tabular}{lccccc}
\hline
Method & PSNR$\uparrow$ & SSIM$\uparrow$ & LPIPS$\downarrow$ & DISTS$\downarrow$ \\
\hline
Traditional & 20.6 & 0.843 & 0.216 & 0.220 \\
ConvBEERS   & 27.5 & 0.932 & 0.068 & 0.077 \\
\hline
\end{tabular}
\end{table}

\begin{table}[t]
\centering
\caption{Physical quality metric on OpenAerialDataset. 
}
\label{tab:runtime_simulated}
\begin{tabular}{lcc}
\hline
Method &  MTF@Nyq & SNR @ $L_0/L_1$ \\
\hline
Traditional & N/A & 65.33 / 138.51  \\
ConvBEERS   & 20\% & 66.27 / 156.07  \\
\hline
\end{tabular}
\end{table}
    
\subsection{Robustness to blur degradation}
The robustness of \mbox{ConvBEERS} to blur degradation is further investigated under increasingly severe imaging conditions. The model is trained on \textit{Sim-Degraded-Variable/Sim-Reference-Fixed} datasets (Tab.~\ref{tab:datasets}) and evaluated on simulated degraded images generated at fixed MTF levels. Three blur levels are considered: low, medium, and high, corresponding to $\mathrm{MTF}_{7\%}$, $\mathrm{MTF}_{5\%}$, and $\mathrm{MTF}_{3\%}$, respectively.
For each input image, the MTF is estimated both before restoration (on degraded input) and after restoration by \mbox{ConvBEERS}. Results (Tab.~\ref{tab:MTF_robustness})  show that the restored MTF is systematically higher than the input MTF across all degradation levels. Importantly, the gain in MTF remains relatively stable as blur severity increases. Even for the most degraded input ($\mathrm{MTF}_{3\%}$), \mbox{ConvBEERS} provides a significant improvement in spatial resolution.

These results indicate that \mbox{ConvBEERS} trained on variable degradation conditions does not overfit to a single nominal operating point, but instead learns a robust restoration behaviour that generalises to unseen blur levels. This robustness is a key property for operational satellite imagery, where imaging conditions may deviate from nominal specifications due to acquisition geometry, temporal variations, or sensor ageing.

\begin{table}[t]
\centering
\caption{MTF before and after restoration with \mbox{ConvBEERS} for different blur levels on simulated degraded data.}
\label{tab:MTF_robustness}
\small
\begin{tabular}{lccc}
\hline
\textbf{Deg. Level} &
\textbf{MTF (Deg.)} &
\textbf{MTF (Rest.)} &
\textbf{$\Delta \mathrm{MTF}$} \\
\hline
MTF$_{7\%}$ \quad & 7.41\% & 20.17\% & +0.1276 \\
MTF$_{5\%}$ \quad & 5.58\% & 15.55\% & +0.0997 \\
MTF$_{3\%}$ \quad & 3.36\% & 13.83\% & +0.1047 \\
\hline
\end{tabular}
\end{table}

    
\subsection{Experiments on Real Pléiades Data}

For real-world validation, we use optical panchromatic imagery acquired by the Pléiades satellite constellation operated by the French Space Agency (Centre National d’Études Spatiales, CNES), which provided access to raw products. In this experiment, we aim to evaluate the generalisation capability of \mbox{ConvBEERS}. Thus, the Pléiades images are not used to train or fine-tune the model. 

Evaluation is conducted through qualitative visual inspection and comparison with the conventional restoration pipeline. In addition, a blind visual assessment was conducted in which observers compared the outputs of both methods without knowledge of the processing approach. Both methods produce visually close results, with comparable levels of sharpness and noise control. However, in a blind evaluation, \mbox{ConvBEERS-restored} images were judged as more visually comfortable, with improved edge definition and no evident non-physical artefacts or excessive ringing.

Overall, these observations indicate that the learning-based restoration model generalises well from simulated data to real Pléiades imagery, while preserving physically plausible image characteristics and offering perceptually favourable rendering.

\begin{figure}[htbp]
    \centering
    
    \begin{subfigure}{0.48\linewidth}
        \centering
        \includegraphics[width=\linewidth]{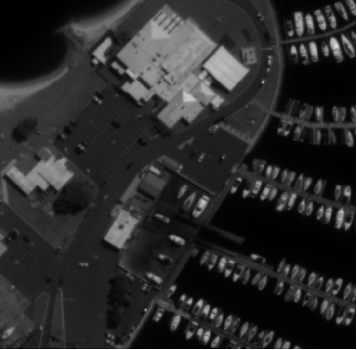}
        \caption{Raw Pleiade}
        \label{fig:sub1}
    \end{subfigure}
    
    \vspace{3mm}
    
    \begin{subfigure}[b]{0.48\linewidth}
        \centering
        \includegraphics[width=\linewidth]{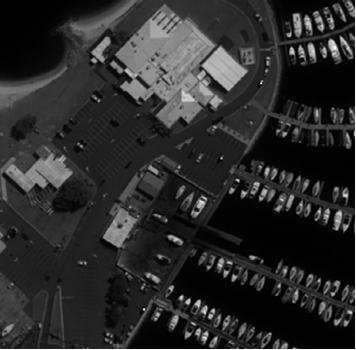}
        \caption{Traditional}
        \label{fig:sub2}
    \end{subfigure}
    \hfill
    \begin{subfigure}[b]{0.48\linewidth}
        \centering
        \includegraphics[width=\linewidth]{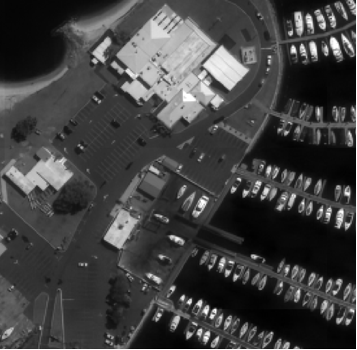}
        \caption{ConvBEERS}
        \label{fig:sub3}
    \end{subfigure}
    \footnotesize Pléiades © CNES 2022-2023, Distributed by CNES PWH, all rights reserved. Commercial use is prohibited.
    
    \caption{Comparison of raw Pléiades image and restored results using the traditional algorithm and ConvBEERS.}

    \label{fig:three_images}
\end{figure}

\subsection{Impact on an application scenario: object detection on the DIOR dataset}

    To demonstrate the added value of learning-based restoration in a fully integrated processing pipeline, we evaluate the impact of \mbox{ConvBEERS} on an object detection task using the DIOR dataset \cite{DIOR}. This well-known dataset in the Earth observation community comprises 23,463 optical remote sensing images covering 20 classes. Object detection is performed using the YOLOv11n model \cite{yolo11}. The nano version, combined with a 256x256 downsampling, ensures that resulting performance reflects realistic edge-system constraints.
    
    \FigDiorPipeline
    
    As displayed in Fig.~\ref{fig:object_detection_pipeline}, the original DIOR images are degraded using the simulation framework described in Sec.~\ref{sec:datasets}, generating pairs of simulated raw and simulated reference images for training. Two test sets with different degradation levels are then created: \textit{Deg-Med}, representative of typical satellite sensor degradation, and \textit{Deg-High}, which applies more severe degradation to further challenge the restoration process. The corresponding degradation parameters are detailed in Tab.~\ref{tab:dior_degradation_parameters}. Then, images are restored using ConvBEERS. Finally, YOLOv11n is trained on all three configuration (reference, degraded and restored). The trained model are used to output detection metrics comparing the baseline (reference images) with both degraded and restored images.

    \begin{table}[t]
        \centering
        \caption{Summary of simulated and real DIOR datasets used in this work.}
        \label{tab:dior_degradation_parameters}
        \footnotesize
        \begin{tabular}{lccc}
        \hline
        \textbf{Dataset} & \textbf{MTF} & \textbf{SNR @ $L_0$ / $L_1$} & \textbf{\#Patches} \\
        \hline
        Deg-Med   & 1--2\%     & 30 $\pm$ 5 / 70 $\pm$ 5 & 18\,000 / 2000 / 3463\\
        Deg-High    & 0.1--0.5\%  & 10 $\pm$ 5 / 50 $\pm$ 5       & 18\,000 / 2000 / 3463\\
        \hline
        \end{tabular}
        \vspace{1mm}
        
        \footnotesize
        $L_0 = 25$~W/m$^2$/sr/$\mu$m and $L_1 = 100$~W/m$^2$/sr/$\mu$m. \#Patches is reported as \textit{train / val / test}. Patches are dowsampled to $256 \times 256$ pixels. 
    \end{table}

    Tab.~\ref{tab:dior_results} displays the results for all three datasets. First, we can observe that the model achieves a strong baseline performance, with 75.90\% mAP@50 and 54.10\% mAP@90. These values indicate that the training process is under control, with no major flaws.

    Regarding the \textit{Deg-Med} experiment, we observe that degradation slightly impacted the detection model's performance, resulting in decreases of 2.6\% for both mAP@50 and mAP@90. However, the restored version \textit{Restored images} almost fully recovered the original performance.

    Regarding the \textit{Deg-High} experiment, the degradation effect is stronger. The raw images suffer a pronounced performance drop, especially in mAP@50 (-8.60\%) and precision (-22.80\%). It indicates that detection is significantly compromised under such heavy degradation. In contrast, the restored images dramatically recover performance, bringing mAP@50 (72.37\%) and precision (85.03\%) closer to reference levels.

    Overall, the results show that even when severe degradation heavily impacts detector performance, using our proposed light restoration model significantly improves performance.

    \begin{table}[htbp]
        \centering
        \caption{Object detection comparison between degraded and restored DIOR datasets with Yolo11n on 256 x 256 pixel images.}
        \label{tab:dior_results}
        \footnotesize
        \setlength{\tabcolsep}{6pt}
        \renewcommand{\arraystretch}{1.15}
        \begin{tabular}{llcccccc}
            \toprule
            \textbf{Dataset} & \textbf{Set} & \textbf{mAP50} & \textbf{mAP90} & \textbf{Prec.} & \textbf{Rec.} \\
            \midrule
            
            Reference & N/A & 75.90\% & 54.10\% & 86.60\% & 69.30\% \\
            
            \midrule
            \multirow{2}{*}{Deg-Med}
                 & Deg. & 73.33\% & 51.5\% & 85.00\% & 66.93\% \\
                 & Rest. & 75.52\% & 53.88\% & 87.47\% & 68.32\% \\
            \midrule
            \multirow{2}{*}{Deg-High}
                 & Deg. & 67.30\% & 46.29\% & 63.80\% & 60.70\% \\
                 & Rest. & 72.37\% & 50.57\% & 85.03\% & 66.36\% \\
            \bottomrule
        \end{tabular}
    \end{table}

    \FigDiorExample

\subsection{Embedded Deployment}


    As a final experiment, we benchmark \mbox{ConvBEERS} against a traditional Pléiades-inspired restoration pipeline \cite{pleiades_restoration} on the Versal Xilinx VCK190 FPGA platform. The Versal has gained traction in the space community, thanks to its radiation-tolerant specifications \cite{versal_heavy_ion, versal_heavy_ion_radecs}, and state-of-the-art performances \cite{versal_plane_detection_yolo, versal_ship_detection_yolo, versal_spaice_project}. In this work, we use a Deep Learning Processor Unit (DPU) \cite{xilinx_dpu_vitisai_documentation, versal_dpu_for_CNN}, designed to accelerate CNN models. 
    \mbox{ConvBEERS} is evaluated on the DPU accelerator. The baseline pipeline is executed only on the CPU because its algorithmic structure is not compatible with the DPU's operator constraints ~\cite{xilinx_dpu_vitisai_documentation}. Since FPGA deployment of AI models typically requires quantisation, \mbox{ConvBEERS} is converted to INT8 precision using post-training quantisation \cite{post_training_quant}. Tab.~\ref{tab:quantification_loss} reports the results of a quantised \mbox{ConvBEERS} compared to its floating-point version as well as the baseline restoration. Firstly, the difference between the outputs of the quantised and floating-point \mbox{ConvBEERS} remains limited, with a Mean Absolute Error of 1.39\%. This result ensures that the restored image is similar before and after quantisation. We observe a modest decrease in PSNR (from 28.6\,dB to 26.7\,7dB). This degradation is expected for a restoration task. Nevertheless, the model still shows strong restoration performance, as reflected by an SSIM value close to 88.5.

       \begin{table}[t]
            \centering
            
            \caption{Quantization performance for a $13000 \times 9000$ panchromatic swath image. }
            \label{tab:quantification_loss}
            \small
            \begin{tabular}{lcccccc}
            \hline
            \textbf{Model} &
            \textbf{Prec.} &
            \textbf{Size (Mb)} &
            \textbf{PSNR (dB)} &
            \textbf{SSIM} \\

            \hline
            Baseline & Float32  & 1.97 & 28.37 & 0.600 \\
            \mbox{ConvBEERS} & Float32 &  4.78 & 28.58 & 0.887 \\
            \mbox{ConvBEERS} & Int8    &  2.01 & 26.77 & 0.848 \\
            
            \hline
            \end{tabular}
            \vspace{1mm}
        
            \footnotesize{The MAE between \mbox{ConvBEERS} Float32 and Int8 outputs is 0.0139, with a Std of 0.0413.}

        \end{table}

        \begin{table}[t]
            \centering
            \caption{Benchmark on Versal AI Edge VCK190 FPGA for a $13600 \times 9016$ panchromatic swath image. \footnotesize{FPS are given for 640x640 patches}}
            \label{tab:fpga_benchmark}
            \scriptsize

            \begin{tabular}{lcccc}
            \hline
            \textbf{Method} &
            \textbf{Latency (s)$\downarrow$} &
            \textbf{FPS}$\uparrow$&
            \textbf{Power (W)$\downarrow$} &
            \textbf{Pxls/s/W$\uparrow$} \\

            \hline
            Baseline (CPU) & 1724 & 0.2 & 22.5 & 3161 \\ 
            \mbox{ConvBEERS} (DPU) & 42 &  7.2 &  29.6 & 98965 \\
            \hline
            \end{tabular}

        \end{table}
    
    Tab.~\ref{tab:fpga_benchmark} reports the deployment results on the Versal VCK190. The traditional baseline is slow on the Versal CPU, with an inference time of  $\sim$28\,min for the whole swath. In contrast, \mbox{ConvBEERS} takes advantage of the DPU accelerator to process the swath in 42\,s, achieving up to 7.2 FPS on $640 \times 640$ patches. Even though the power consumption increases with the use of the DPU (from 22.5\,W to 29.6\,W), results show that this approach is compatible with onboard real-time processing.

\section{Discussions}

From an operational perspective, the results highlight the relevance of lightweight learning-based restoration models for both ground-based and onboard processing. Although traditional restoration pipelines remain robust, their sequential structure and computational cost limit their suitability for resource-constrained environments. In contrast, the proposed \mbox{ConvBEERS} approach achieves comparable image quality while significantly reducing processing time, making it better aligned with near-real-time operational constraints. 

Beyond the comparison with traditional pipelines, \mbox{ConvBEERS's} robustness to variations in blur and noise levels is a key advantage in practical scenarios, where imaging conditions may deviate from nominal specifications due to acquisition geometry, temporal effects, or sensor ageing. Learning-based restoration enables earlier and more flexible exploitation of satellite imagery.

When combined with onboard object-detection tasks, \mbox{ConvBEERS} significantly improved the AI model's performance. These results, coupled with ConvBEERS' ability to preserve the physical plausibility of the image, reinforce the suitability of learning-based restoration as a task-oriented preprocessing step for onboard AI applications in remote sensing. 

Nevertheless, the \mbox{ConvBEERS} model currently only tackles radiometric degradation, lacking geometric enhancement capabilities. Thus, in cases of geometrical degradation, such as a significant band misalignment, coupling \mbox{ConvBEERS} with another algorithm might be needed.

Overall, the reasonable overhead in power, coupled with the growing performance of satellite's computing capability, converge towards the use of ConvBEERS-like architecture for future spaceborn processing.

\section{Conclusion}
\label{sec:conclusion}

In this paper, we valuated the relevance of learning-based methods for onboard preprocessing of raw satellite imagery. The residual layers of the EDSR model are adapted into our proposed restoration model, \mbox{ConvBEERS}. The proposed model was first evaluated using full-reference image quality metrics (SSIM, PSNR, LPIPS, and DISTS) and physical image quality metrics (SNR, MTF). Results on OpenAerialMap images and real Raw/CNES-L1 pairs indicate that \mbox{ConvBEERS-restored} images exhibit characteristics comparable to those obtained with traditional restoration pipelines. When evaluated across varying levels of blur degradation, \mbox{ConvBEERS} demonstrated strong robustness, yielding stable restored MTF values across degradation scenarios. Furthermore, \mbox{ConvBEERS} was assessed as a lightweight preprocessing module for AI-based object detection. Experimental results show that detection performance consistently improves, confirming the benefit of restoration for onboard AI tasks. Finally, onboard deployment confirms its suitability for spaceborne applications, demonstrating significant acceleration while maintaining restoration quality.

In future work, the \mbox{ConvBEERS} architecture could be improved with remote sensing imagery knowledge. Indeed, because remote sensing data differ from natural images, incorporating physical priors \cite{PINN_for_computer_vision} may prevent the emergence of non-physical geometric artifacts.
Another promising direction would be to integrate restoration directly within a task-dependent AI model, simplifying the onboard processing pipeline by eliminating the need for two independent models.
Finally, while this work focuses on radiometric restoration of raw images, it would be very interesting to couple with learning-based geometric restoration techniques, such as the one proposed by Del Prete et al. in PyRawS \cite{pyraws_thraws}. The result would be a fully operational restoration model that can work with any raw image.

{
    \small
    \bibliographystyle{ieeenat_fullname}
    \bibliography{main}
}


\end{document}